\documentclass[conference]{IEEEtran}
\usepackage[utf8]{inputenc}
\usepackage[square,sort,comma,numbers]{natbib}
\usepackage{multicol}
\usepackage{graphicx}
\usepackage{todonotes}
\usepackage{xcolor}
\usepackage{comment}
\usepackage{varwidth}
\usepackage{enumitem}
\usepackage{zref-xr}
\usepackage{caption}
\usepackage{multirow}
\usepackage{amsmath}
\usepackage{amssymb}
\usepackage{subfigure}
\usepackage{algorithm}
\usepackage{algorithmic}
\IEEEoverridecommandlockouts
\title{Analyzing Worldwide Social Distancing through Large-Scale Computer Vision}

\author{
    \IEEEauthorblockN{Isha Ghodgaonkar, Subhankar Chakraborty\IEEEauthorrefmark{2}, Vishnu Banna, Shane Allcroft, Mohammed Metwaly, Fischer Bordwell,  \\
    Kohsuke Kimura, Xinxin Zhao, Abhinav Goel, Caleb Tung, Akhil Chinnakotla, Minghao Xue \\
    Yung-Hsiang Lu, Mark Daniel Ward, Wei Zakharov, David S. Ebert, David M. Barbarash, George K. Thiruvathukal\IEEEauthorrefmark{3}}
    \IEEEauthorblockA{Purdue University, West Lafayette, IN, USA\\}
    \IEEEauthorblockA{\IEEEauthorrefmark{2}Indian Institute of Technology Madras, Chennai, India\\}
    \IEEEauthorblockA{\IEEEauthorrefmark{3}Loyola University Chicago, IL, USA\\}
}

\begin{document}
\maketitle

\begin{abstract}
In order to contain the COVID-19 pandemic, countries around the world have introduced social distancing guidelines as public health interventions to reduce the spread of the disease. However, monitoring the efficacy of these guidelines at a large scale (nationwide or worldwide) is difficult. To make matters worse, traditional observational methods such as in-person reporting is dangerous because observers may risk infection. A better solution is to observe activities through network cameras; this approach is scalable and observers can stay in safe locations. This research team has created methods that can discover thousands of network cameras worldwide, retrieve data from the cameras, analyze the data, and report the sizes of crowds as different countries issued and lifted restrictions (also called ``lockdown''). We discover
11,140 network cameras that provide real-time data and we present the results across 15 countries. We collect data from these cameras beginning April 2020 at approximately 0.5TB per week. After analyzing 10,424,459 images from still image cameras and frames extracted periodically from video, the data reveals that the residents in some countries exhibited more activity (judged by numbers of people and vehicles) after the restrictions were lifted. In other countries, the amounts of activities showed no obvious changes during the restrictions and after the restrictions were lifted. The data further reveals whether people stay ``social distancing'', at least 6 feet apart. This study discerns whether social distancing is being followed in several types of locations and geographical locations worldwide and serve as an early indicator whether another wave of infections is likely to occur soon. 

\end{abstract}

\vspace{-5pt}
\section{Introduction}
On November 11, 2019, the first known case of COVID-19 was reported~\cite{ma_chinas_2020}.  In January 2020, the World Health Organization (WHO) warned that the fast-spreading virus could reach other parts of the globe~\cite{nebehay_who_2020}. By March 2020, COVID-19 cases had spread around the world, and the disease had been officially labelled as a pandemic~\cite{saavedra_global_2020}. Since then, countries have implemented different policies to slow down the spread of the disease, for example, stopping non-essential activities (also called ``lockdown'' policies), encouraging people to follow social distancing guidelines, and more recently encouraging or requiring wearing face covers. In public health, social distancing is defined as maintaining a safe distance to reduce close contact between people. This distance is generally around 6 feet. 

\begin{figure}
  \centering
    \includegraphics[width=0.45\textwidth]{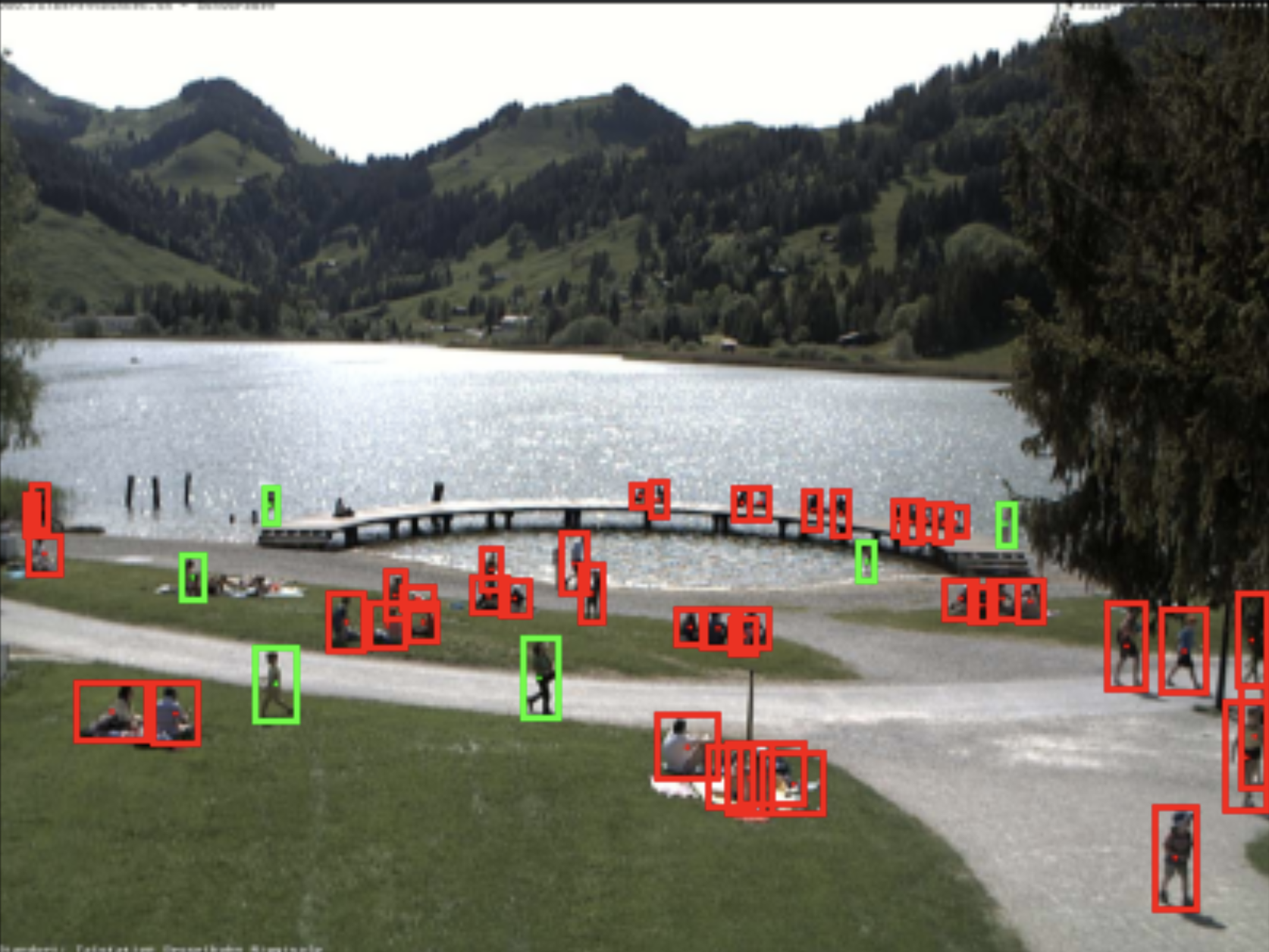}
  \caption{A park in Schwarzsee, Switzerland, on May 22, 2020. Red and green bounding boxes represent people violating social distancing guidelines and following them, respectively, without consideration for families traveling in groups.
  Image source: Snoweye.com}
  \label{fig:schwarzsee}
\end{figure}

Observing the effects of social distancing is difficult with traditional methods of journalism. Dispatching reporters and researchers to a location could expose them to health risks. Some countries have strict travel restrictions and sending reporters is not possible. Other methods rely on self reporting; such methods have well-known biases as they capture only the people that are willing to report their locations. Moreover, it is difficult for a single person or any organization to collect enough data at the same locations every day during a month-long lockdown and observe the trend. Any method that involves human labor is not scalable to nationwide or worldwide observation. Therefore, it is necessary to develop methods of observation without the  physical presence of humans.

Many cities deploy traffic cameras and make the data available to the general public. We hypothesize that if fewer people are leaving their homes and travelling to places, there must be fewer vehicles on highways or city streets. Thus, the number of vehicles can provide an indirect indication of whether people are following social distancing guidelines.  

This research team uses the globally deployed network cameras to observe activities during and after lockdown. This study is built on the team's existing work of discovering network cameras~\cite{LU2019safetycamera, LU2019systemmethod, LUDailey2020discovery} and retrieving and analyzing the data~\cite{see_world_through_cameras}.  
Thousands of network cameras have already been deployed worldwide for monitoring traffic, observing national parks and other locations. This study observes responses to pandemic intervention policies in 63 countries utilizing 11,140 cameras. 
This study focuses on 
observing trends in selected location archetypes (such as `street corner' or `restaurant').
The study uses computer vision to select this subset of cameras out of 30,702 originally discovered cameras by removing frozen camera streams and the cameras that provide no information about human activities (a planetary camera pointed at the night sky, for instance). 
Specifically, the study identifies 9,752 cameras to analyze traffic and 3,254 cameras to count people. 
This study collects and analyzes data from April 1, 2020, to August 1, 2020. The analysis includes 15 selected countries, 2,973 cities, 10,424,459 images extracted from still image cameras and video segments, 301,398 detected people, and 12,459,271 detected vehicles to observe potential changes in crowd sizes,  vehicle presence, or both. This paper provides insight about the ability of computer vision to analyze the vast, diverse visual data available from network cameras. 
Overall, object detection performs well enough to give reasonable estimates of the numbers of people and vehicles in data.

The analysis shows that many countries follow the lockdown: the numbers of people and vehicles increased noticeably after the policies were lifted.
Some areas do not show a clear pattern over the four months.
\vspace{1pt}

This paper has the following major contributions:
\begin{enumerate}
\item We believe this is the first study that utilizes the data from network cameras to observe human activities during and after lockdown in response to COVID-19. Moreover, we are not aware of any other study comparable to this in terms of scale.

    \item The team creates computer tools that can collect visual data  periodically from the thousands of network cameras.
    \item This paper develops a method to strategically narrow down over 10 TB of image and video data from 30,702 cameras to less than half to use for analysis.
    \item This paper presents the patterns in crowd sizes and traffic over time in 15 countries.
    \item This paper evaluates existing computer vision methods including crowd counting and object detection on a representative sample of network camera data and discovers the extent to which computer vision can perform successfully on the data.
    \item This paper develops the basis for an algorithm to detect social distancing in images. This paper uses the algorithm to discover geographical locations and types of locations in which there is a lack of social distancing.
\end{enumerate}
\vspace{-2pt}
\section{Related Work}
This section considers related work in three categories:  1) detecting people and vehicles in images and video, 2)  estimating social distancing using computer vision, and 3) analyzing social distancing using other digital means.

\subsection{Detecting people and vehicles in images}

 Crowd counting is a specialized class of techniques to estimate the number of people in an image or a video. The three broad approaches to crowd counting are via detection, regression and density estimation. Early attempts \cite{zhao2003bayesian,chen2012feature} at crowd counting involved the use of detection and regression. Detection methods include You Only Look Once (YOLO) \cite{redmon2016you} and Single-Shot-Detector (SSD) \cite{liu2016ssd}. Object counting may use hierarchical object counting \cite{Goel_2020}. These early methods are inaccurate with estimating density in moderate to dense crowds. Many methods \cite{liu2019crowd,shi2019counting,wang2019learning,zou2020crowd} for popular crowd counting datasets \cite{zhang2016single,idrees2018composition} use deep learning to estimate crowd density maps and integrate over the maps to predict the counts. Many crowd counting methods fail to perform well on  data from network cameras due to the assumption of the presence of dense crowds. In data with sparse or no crowds, the methods are unreliable.

 Vision methods use the motion of a vehicle to separate these from  background and are hence unsuitable on still images. These methods can be broadly divided into three types: \cite{abdulrahim2016traffic}: 1) using background subtraction \cite{manikandan2013video}, 2) using continuous video frame difference \cite{li2011vehicles}, and 3) using optical flow \cite{liu2013optical}. YOLO along with road pixel segmentation \cite{alvarez2012road} is used in some methods \cite{song2019vision} for vehicle detection and tracking. Workman et al. \cite{workman2020dynamic} use overhead imagery to understand patterns in traffic flow.
Open-source software \cite{mmdetection,darknet13,hasan2020pedestrian,glenn_jocher_2020_3785397} facilitates the process of training object detectors on custom data sets or using pre-trained models on common datasets \cite{lin2014microsoft,shao2018crowdhuman}. In this work we find that robust object detection models perform best on our data. 

\subsection{Estimating social distancing using AI}
 Existing work explores how wireless and networking technologies combined with AI can help enable and enforce social distancing \cite{nguyen2020enabling}. Visual Social Distancing \cite{cristani2020visual} introduces a method to estimate interpersonal distance given a 2D image. This method is based on pose estimation and requires high resolution images with relatively close-up people. In many images obtained from network cameras, people are too pixelated for this method to perform successfully.  Yang et. al \cite{yang2020vision} and Khandelwal et. al \cite{kh2020using} utilize perspective transformation to change a given image's perspective to a bird's eye view before estimating social distancing. The transformation requires human effort and therefore is not scalable. Several open-source methods \cite{senaratne} use the centroids of detected people's bounding boxes in order to estimate distance between people. These methods do not consider depth differences. Disnet \cite{haseeb2018disnet} provides a deep learned model to estimate distance from the camera, but is not lightweight and therefore not suitable for our use case given the size of our data. To the best of our knowledge, there is no method to estimate social distancing in heterogeneous camera data that is lightweight (and therefore scalable to data of this size), automatic, robust to perspective change and depth in 2D images.

\subsection{Analyzing worldwide responses}
Several studies analyze adherence to social distancing interventions using data-driven methods. For example, Google Cloud Mobility Reports is used to gauge behavioral responses to social distancing policies internationally \cite{mobilitychanges, van2020not}. This data may be biased because users opt in to save their location history. Similarly, 
mobility from wearable trackers or mobile phones can be used \cite{wearabletracker}
 \cite{oliver2020mobile}. As far as we know, there is no study that analyzes social distancing using camera data to the scale presented in this paper.

\section{Camera Discovery and Data Collection}
The following sections will describe the process to discover network cameras, collect data, manage the data,  narrow down the data for analysis, analyze the data, and present results.
The discovery method finds 30,702 cameras distributed worldwide. Then, we select the cameras based on the data quality in order to perform analysis. The large number of cameras
produces a vast amount of data. This paper discusses the methods to manage over 10 TB of data and select which data to analyze. We then describe the process of choosing state-of-the-art computer vision methods, and show successes and failures of the chosen methods. Finally, we present the trends over time in selected countries.
An increasingly large number of network cameras are available publicly online. Hosted by governments, schools, and private entities, these cameras are sources of live visual data. To observe worldwide responses to COVID-19, thousands of cameras are needed. However, no simple method exists to collect this data. Due to the heterogeneity of website designs, developing automated solutions for network camera discovery is needed.

\subsection{Camera Discovery}
\begin{figure}[htb!]
     \centering
     \includegraphics[width=3.2in]{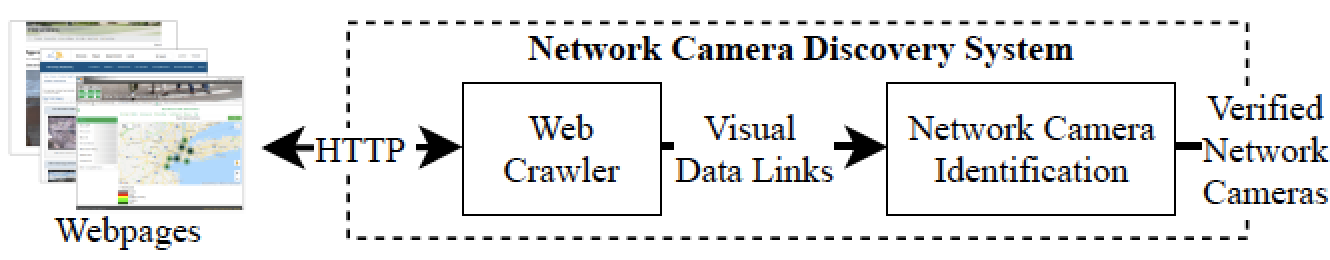}
     \caption{Automatic Network Camera Discovery System \cite{automated_discovery}. The system has 2 main parts: (1) Web Crawler module and (2) Network Camera Identification module. }
     \label{fig:Ryan_system}
 \end{figure}

This team has developed automated solutions to discover network cameras \cite{LUDailey2020discovery}. The process is composed of two modules, as shown in Figure~\ref{fig:Ryan_system}. The first module discovers websites that may contain live visual data. This is achieved by a web crawler that receives a set of seed URLs and tabulates a list of the links found from crawling the seed URLs. The module identifies the links that may contain live data and sends the URLs to the second module.
The module  determines whether the visual data is live or not by checking whether the data changes over time. This process discovers 30,261 live network cameras providing snapshots and an additional 441 live video cameras.

After the web crawler discovers a URL, the crawler parses the website for image formats such as JPEG and PNG. For any visual data link found on a web page, the web crawler module downloads and parses the HTML. The pages are displayed in a web browser environment \cite{automated_discovery} to ensure all web assets are properly loaded to avoid any potential losses of information. The crawler then parses the HTML response and searches for data links common to network cameras such as image-specific visual links (e.g. baseURL /camera id.jpg) and video stream links (e.g. starting with rtmp::// and rtsp::// or ending with .mjpg). 
After the web crawler aggregates potential links for image data, the network camera identification module determines whether such data links connect to network cameras that update the data frequently. The identification distinguishes between active camera data (frequently changing) and web assets (rarely changing). The module retrieves several images from the data links at different times, and after each retrieval, compares the images to determine the change or lack thereof. The module uses three different comparison methods: (1) checksum: compare the file checksum of the images, (2) percent difference: compare the percentage of pixels changed between images, and (3) luminance difference: compare the mean pixel luminance change between images. If the images change over time, the link is considered as connected to a live network camera.

\begin{figure}
    \includegraphics[width=3.2in]{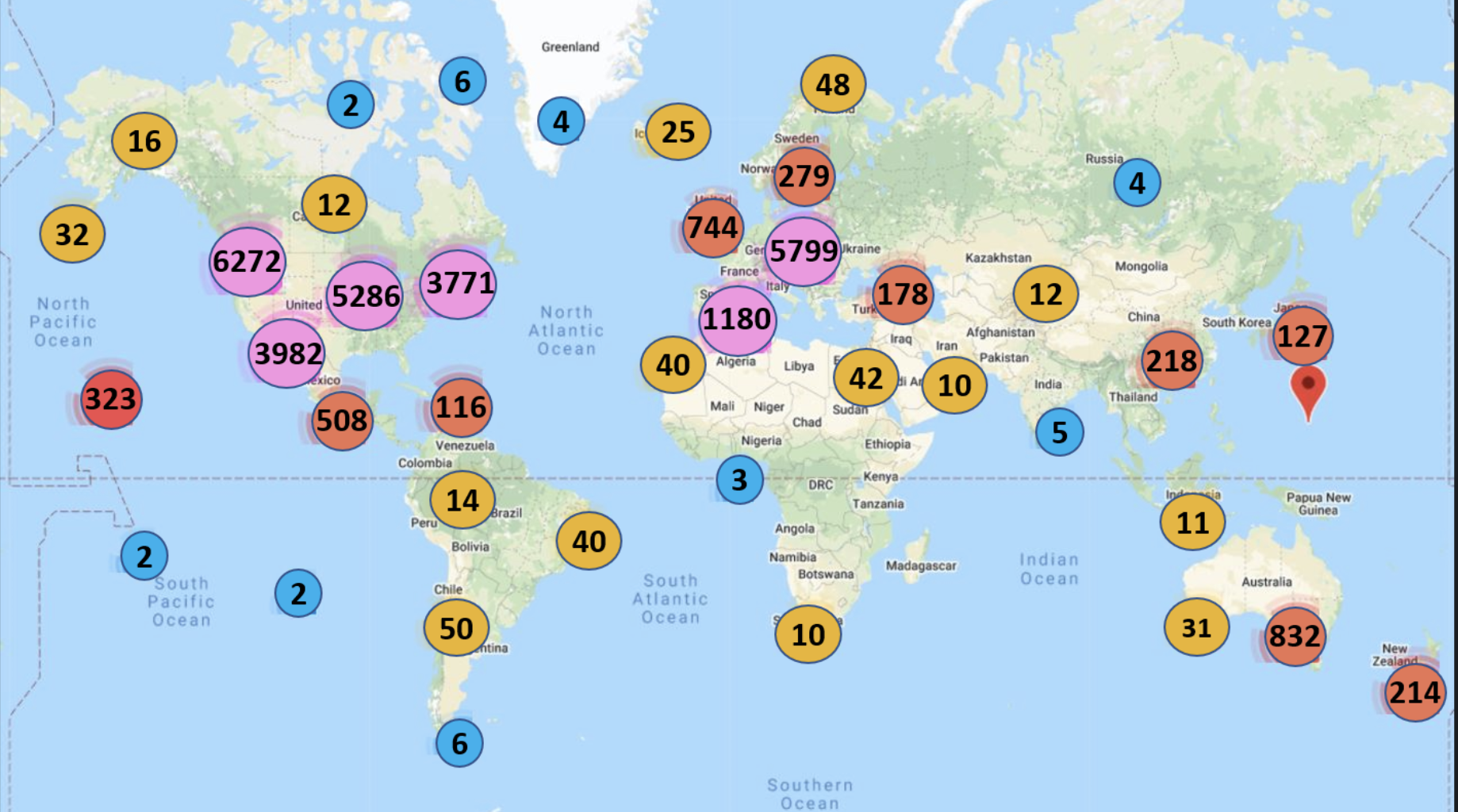}
    \caption{Camera counts of 30,261 discovered cameras 
    \\
    distributed by geographic location.}
    \vspace{0.5cm}
    \includegraphics[width=3.2in]{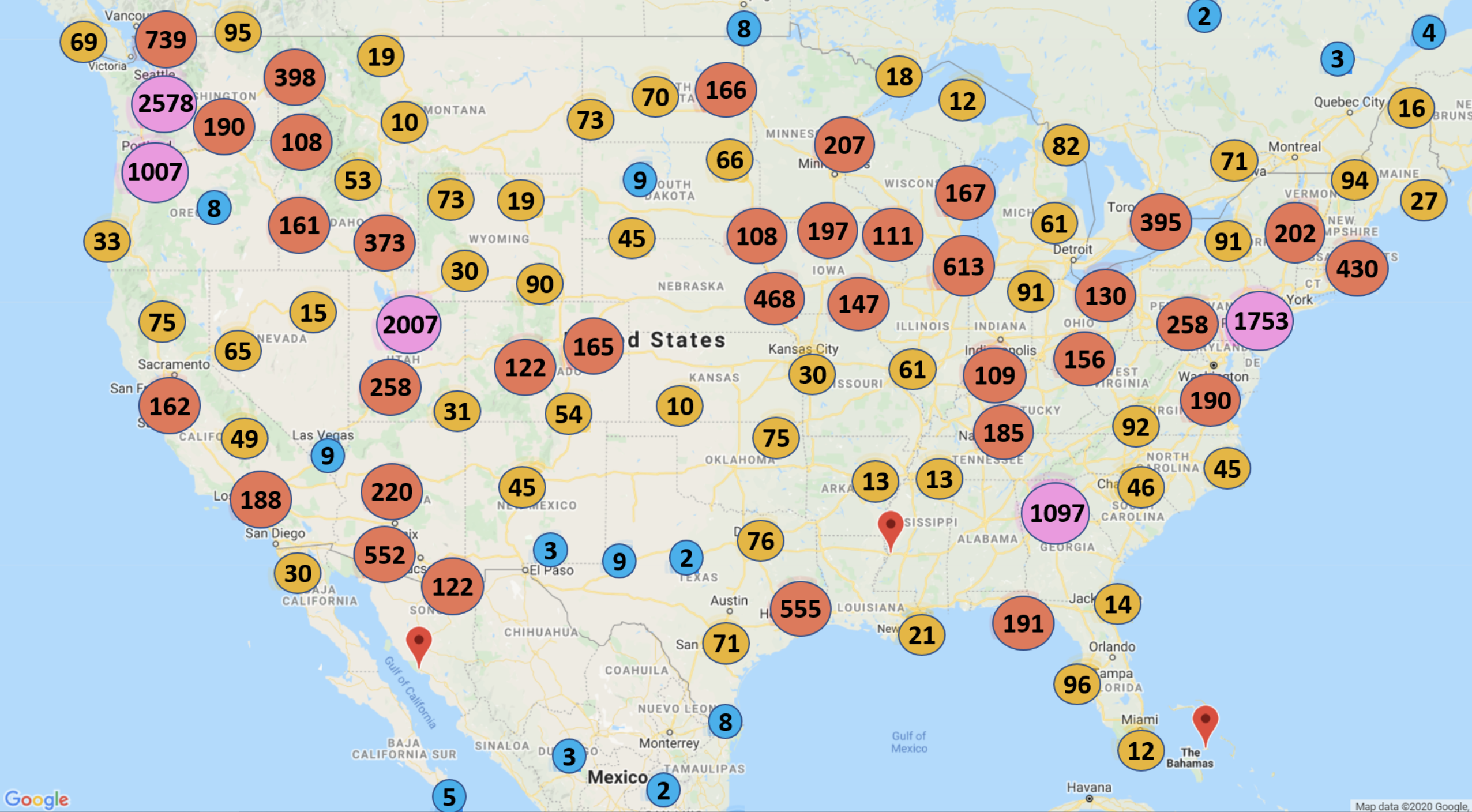}
    \caption{USA map of discovered cameras.}
    \label{fig:map}
    
\end{figure}

\subsection{Data Collection}
After the network cameras are discovered, a computer program runs at defined time intervals and submits jobs to save data from subsets of cameras. The intervals begin on varying time stamps each day to ensure data is collected regularly each week for every camera. On average, we collect 5 images or video clips per day per camera. The image archiver program is capable of handling several image data formats. Video data is downloaded using one of several APIs built upon a plugin system for developers to access video live stream data. The compatible plugin link for each video link is obtained before the relevant API methods are invoked to download data. 
    
\vspace{-1mm}

\section{Data Management at Scale}
Capturing visual data from 30,702 cameras and running object detection methods on a significantly large subset of these cameras poses two distinct challenges: 1) obtaining enough computational power to collect visual data and run models on selected data, and 2) running retrieving programs in parallel to ensure that multiple cameras can be obtained simultaneously and indefinitely. 

We address scalability, parallelizability, and compute power in this work by using Cooley, a mid-sized powerful computational cluster at Argonne Leadership Computing Facility. Through the Director's Discretionary program, we possess access to over 3000 node hours per month and 300 TB of persistent storage. Cooley is primarily targeted to high-performance data visualization workloads, which makes it ideal for capturing video and performing analysis. We retrieve data at the aforementioned intervals from all cameras via a set of short jobs that run on Cooley. 

A notable feature of our high-performance computing environment is the enormous amount of networked storage available to store datasets. The image archiver mentioned in section III is designed to utilize all available cores to fetch data simultaneously. As more still image cameras and video cameras are discovered, Cooley will allow seamless expansion. At the time of writing, we possess over 9 TB of image data and over 1 TB of video data, and over 2 PB of storage is available to cluster users.

Cooley contains 126 compute nodes. Each node has two 2.4 GHz Intel Haswell E5-2620 v3 processors (6 cores per CPU, 12 cores total), with 384GB RAM. The system uses FDR Infiniband interconnect. Each compute node contains an NVIDIA Tesla K80 dual-GPU card. We utilize both Cooley and an additional NVIDIA Titan X GPU to run video preprocessing object detection models on selected data.

\vspace{-1mm}

\section{Data Selection and Preprocessing}
It is unnecessary to analyze all images from all 30,261 still image cameras. Therefore, we strategically choose cameras to analyze based on two criteria: camera functionality and scene classification. We attempt to analyze all 441  video live streams.

\subsection{Filtering based on camera functionality}
\vspace{-1mm}
 To avoid analyzing frozen camera streams, we check 4 equally spaced images from the set of saved images of each camera. If all images are identical pixelwise, we disregard this camera stream as it is likely to be frozen. 

\vspace{-1mm}

\subsection{Scene classification}
\vspace{-1mm}
Places365 \cite{zhou2014learning, zhou2017places} is a scene-centric database containing images labeled with one of 365 category labels. This work uses a ResNet classifier trained on Places365 to identify which ``scene", or type of place, a given camera shows.

Due to the uncertainty of classifications, 5 random images are selected from each camera and assigned a scene classification. We use any one of the 5 classifications to determine candidates for analysis. For example, a camera classified as ``highway'' may also pan and zoom to additionally show a ``crosswalk'', and therefore can be used to detect both people and cars.
 We count vehicles from cameras with a ``highway" or ``road" classification. In total, we analyze 11,186 cameras for vehicles. A total of 9,751 cameras from these are functional and found to contain traffic data for at least one day during the 4 months. We search for people in cameras in 23 scene classifications.
 In total, we choose 4,615 cameras to analyze for people, including 441 video streams. From these, 164 video streams and 3,090 image cameras are found to contain people on at least one day out of the 4 months analyzed. In total, we utilize 11,140 cameras in this analysis, of which 1,865 overlap. 

 To determine a singular classification for a camera for statistical purposes, we take the mode of the 5 classifications determined by the model. If there are multiple modes, we select the mode with the highest confidence.  Our cameras span 274 of 365 scene categories, revealing the diversity of our camera database. The scene ``highway" is the most populous category, containing 5,875 cameras, followed by ``desert road" with 2,568 cameras.

\subsection{Preprocessing video data}
\vspace{-1mm}
Each clip for a video camera consists of a one minute video, subject to small variations. Since people move slowly within a video clip, we do not need to detect people in every frame. We pre-process the videos to save every 30th frame. Then, we take the maximum number of people detected amongst these frames as the person count for this clip.

\section{Analysis Methods}
Once we identify which cameras to analyze, we apply computer vision to detect people and vehicles by using two object detection models. The following sections provide details about the object detection models used. It is important to note that while the models may not detect every car on the highway or every person in an image at a very small scale, the object detector is consistent enough for purposes of reporting trends over time.

\subsection{Detecting people in images}
We use a object detection model from Pedestron \cite{hasan2020pedestrian}, specifically, Cascade-HRNet \cite{SunXLW19,WangSCJDZLMTWLX19} trained on the CrowdHuman dataset \cite{shao2018crowdhuman}, to detect people. This model was validated by the developers of Pedestron with a mean average precision (mAP) of 84.1 on the CrowdHuman validation dataset of 4730 images. We additionally validate this model on a validation dataset of 1000 images constructed from our network camera data by labeling ground truth bounding boxes, using an IOU threshold of 0.5, resulting in an F1 score of 0.712 and a mAP of 61.3. We utilize a confidence threshold of 0.3 for inference.

\vspace{-1pt}

\subsection{Detecting vehicles in images}
 We use a YOLOv3 model trained on the MS COCO dataset \cite{lin2014microsoft} to detect vehicles. The model was validated by Ultralytics \cite{glenn_jocher_2020_3785397} on the MS COCO validation set with an mAP of 57.9. MS COCO contains 20 classes, four of which are `car', `truck', `motorcycle', and `bus', which we consider as vehicles. We additionally validate this model on a validation dataset of 1000 images constructed from our network camera data by labeling ground truth bounding boxes, using an IOU threshold of 0.5, resulting in an F1 score of 0.619 and a mAP of 45.6. We utilize a confidence threshold of 0.3 for inference.

 \vspace{-1pt}

 \subsection{Generating time-series plots from the data}
 We run the object detectors on all image data, taking the maximum count as the people count for a given day for a given camera. This accounts for  measurements taken at night, which may have little to no detection results and thereby serves as a smoothing filter on the number of detected people. We then plot the sum of the maximum of the numbers of detected people over all cameras in a country or city to generate graphs of people and vehicles observed over time. Therefore, each point on the graph is composed of data from many cameras rather than a single camera.
 As some cameras have missing data for certain dates during the timeframe analyzed, taking the maximum number of people observed allows us to observe a clearer trend. While this process may keep outliers in the data, we consider these useful. A spike in the data points to a possible scenario warranting investigation. These plots are shown in Figure~\ref{fig:scatter_countries}.
  
\subsection{Total amount of data analyzed}
In total, we analyze 2,498,920 images from still image cameras for people, and 6,174,806 images for vehicles. From 164 video cameras, we analyze 1,750,733 frames for people. In total, we analyze 10,424,459 images over a 4 month timeframe.

\subsection{Calculating Social Distancing}

We develop a method to count the number of social distancing violations in an image based on detected objects. Violations are an important metric that is missed when only counting numbers of people. To count the total number of violations in an image, we check all pairwise combinations of bounding boxes to decide whether each pair of boxes is in violation. 

In order to develop a method that is lightweight, automatic, and robust to depth difference and camera angle, we make the following assumptions:

\begin{enumerate}
    \item All people in the image are the same height of 5.4 feet, which is the height of an average human. 
    \item All detections reported by the object detector are people, including false positives. 
    \item All people in the image are standing upright and bounding boxes have the same aspect ratio.
\end{enumerate}

While we recognize these assumptions decrease the confidence of the reported number of violations, the method works well enough to seek out locations with high levels of violations. The success of this method depends on the accuracy of the object detection model used. In our case, the object detection model is reasonably accurate, and in the majority of cases, the camera is far enough from people to render relative height differences inconsequential. Most people are standing upright in the images, and in cases where people are sitting down, our method still detects violations well as shown in Figure~\ref{fig:schwarzsee}.

\begin{figure}
    \includegraphics[width=0.5\textwidth]{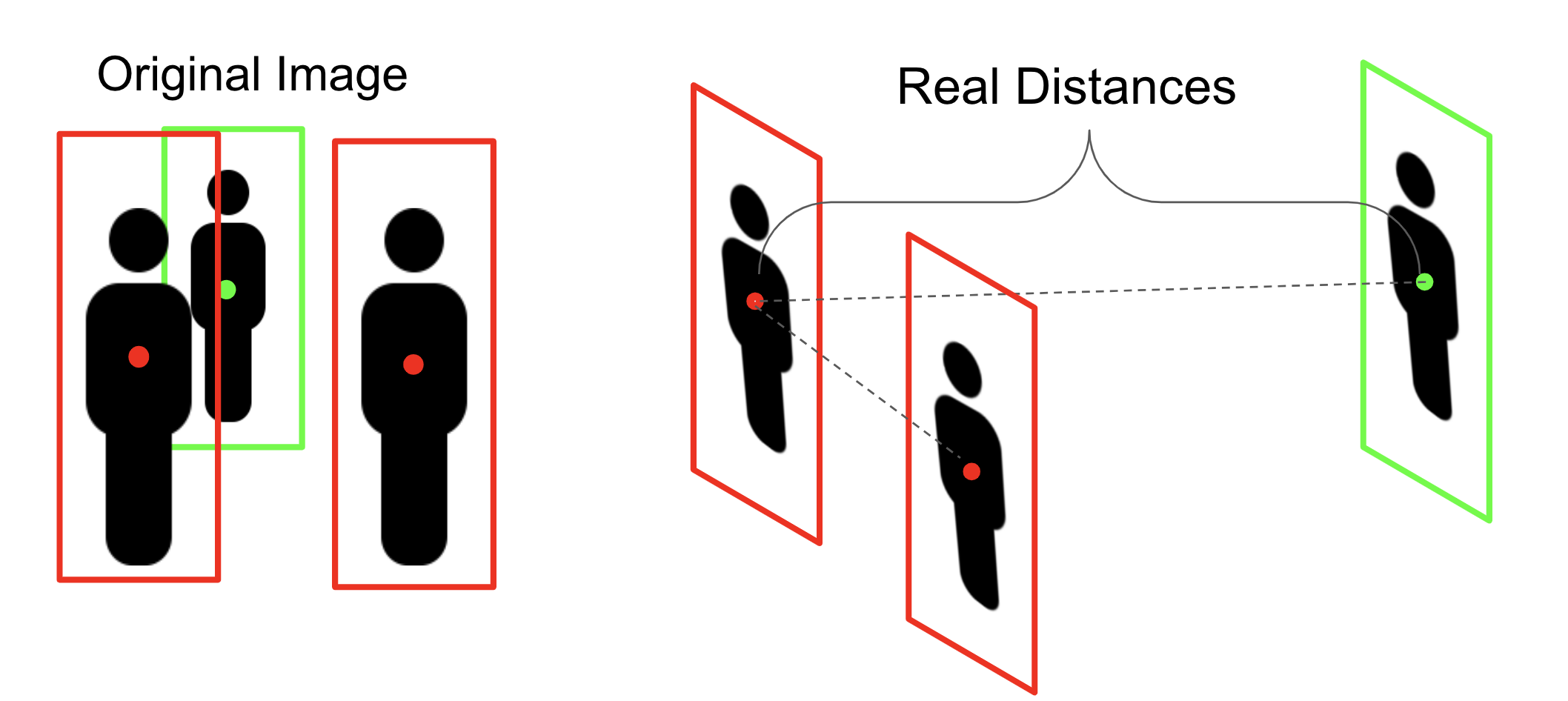}
    \caption{Visualization of social distancing algorithm. Left: How people appear in a 2D image. Right: How the people most likely appear in reality. The pixel distance between the two red boxes in the original image is greater than the pixel distance between the green and red boxes. However, only the red boxes are in violation.}
    \label{fig:distancing}
\end{figure}

\begin{figure}
    \includegraphics[width=0.5\textwidth]{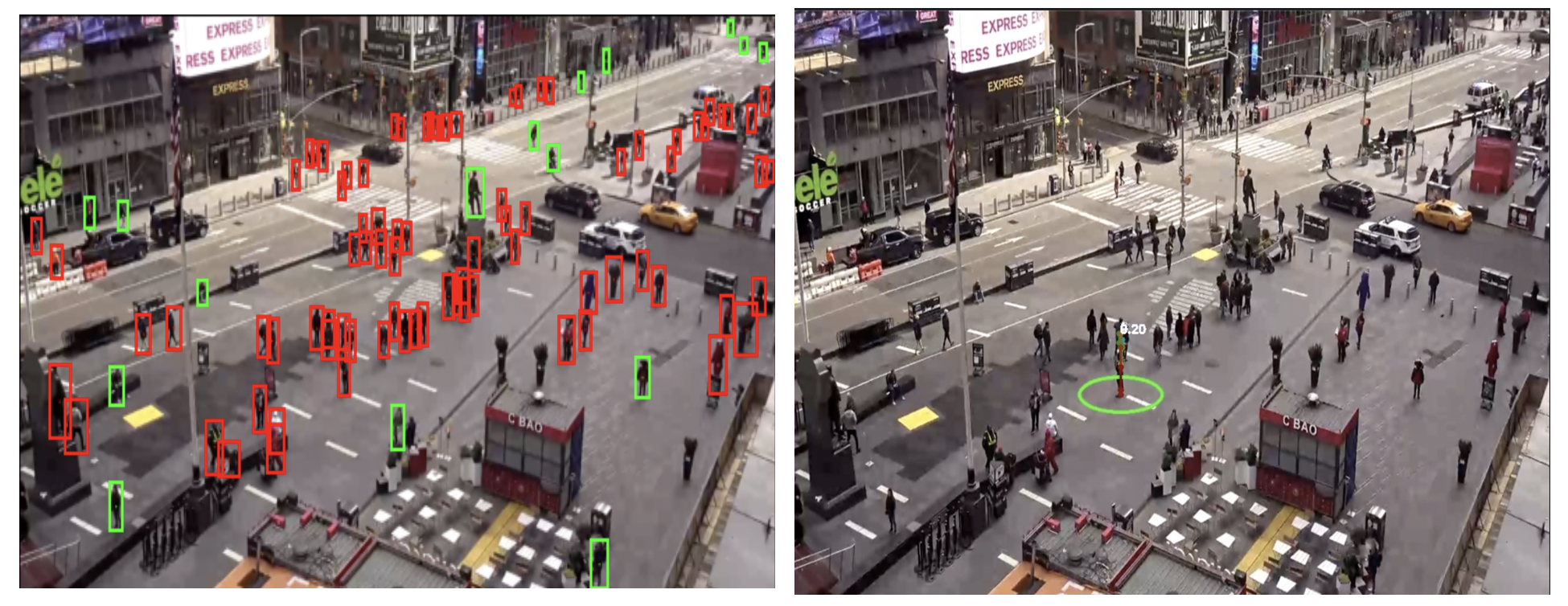}
    \caption{A comparison between the proposed method and Visual Social Distancing \cite{cristani2020visual}. Left: The team’s methods can successfully estimate people that are within six feet. Right: Visual Social Distancing fails to detect most people and violations.  Image source: Earthcam.com}
    \label{fig:distancing2}
\end{figure}

The method works as follows: first, we account for the depth difference between the two bounding boxes. Assuming that all people are of average and equal height allows us to utilize the size of the bounding box as an indication of how far the person is from the camera. We calculate a metric ``depth similarity", $P$, by dividing the area of the smaller bounding box by the area of the larger bounding box. The depth similarity metric approaches 1 as the two bounding boxes approach equal areas and approaches zero as the two people stand at increasingly different distances from the camera. Then, we calculate euclidean distance between the bounding box centers, $D$, measured in pixels. We scale this distance by the average bounding box height to obtain ``inverse relative distance", $ID$. Finally, we multiply $ID$ by $P$ to obtain a number ranging from (0, image height in pixels) unless $D$ is zero, in which case the centers of each bounding box overlap. We do not consider this corner case because empirical evidence shows that this case is unlikely to present itself in realistic data. If the product is greater than 1, this indicates the two people are within an average person's height of one another. However, if this product is greater than (6/5.4) = 1.11, this indicates the two people are within 6 feet of each other. Since the accepted standard of social distancing is normally 6 feet, we use this threshold for our experiments. 
Figure~\ref{fig:distancing2} shows the results of our method as compared to a recent method in literature, Visual Social Distancing \cite{cristani2020visual} that aims to estimate social distancing violations based on a 6 foot radius around a person, but fails to detect many violations. It is important to note, however, that violating social distancing is expected among family members or roommates.

\vspace{-1mm}

\begin{figure}
    \subfigure[]{
    \includegraphics[width=0.22\textwidth]{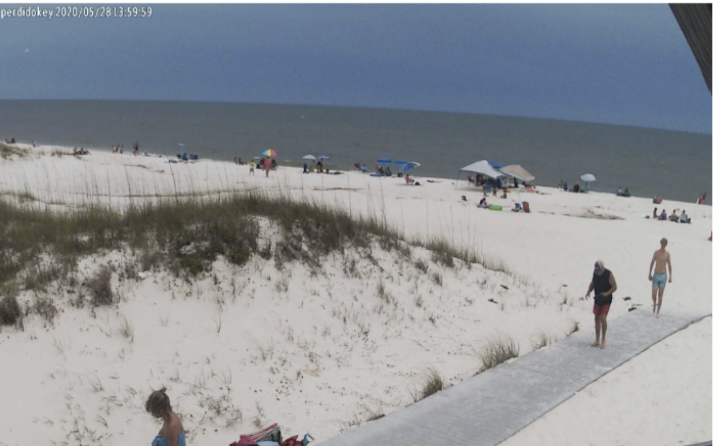}}
    \subfigure[]{
    \includegraphics[width=0.22\textwidth]{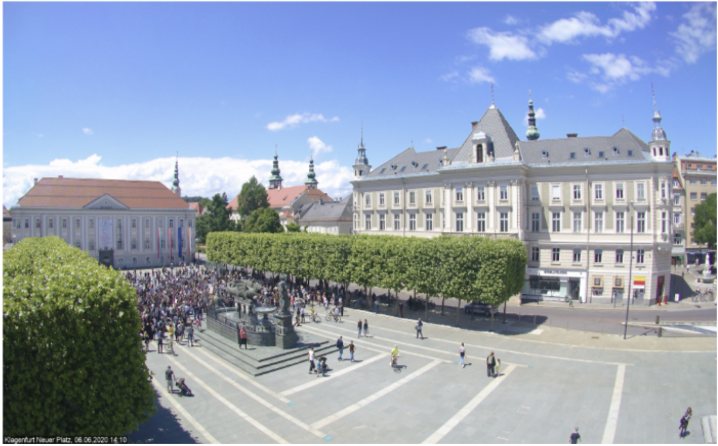}}
    \caption{Successes of the chosen object detection method on selected images. (a) 35 people detected in Ocean Springs, Mississippi, on May 28, 2020. (b) More than 80 people detected in a protest in Klagenfurt, Austria, June 6, 2020. Image sources: Wunderground.com, klagenfurt.it-wms.com}
    \label{fig:successes}
\end{figure}

\begin{figure}
    \subfigure[]{
    \includegraphics[width=0.22\textwidth]{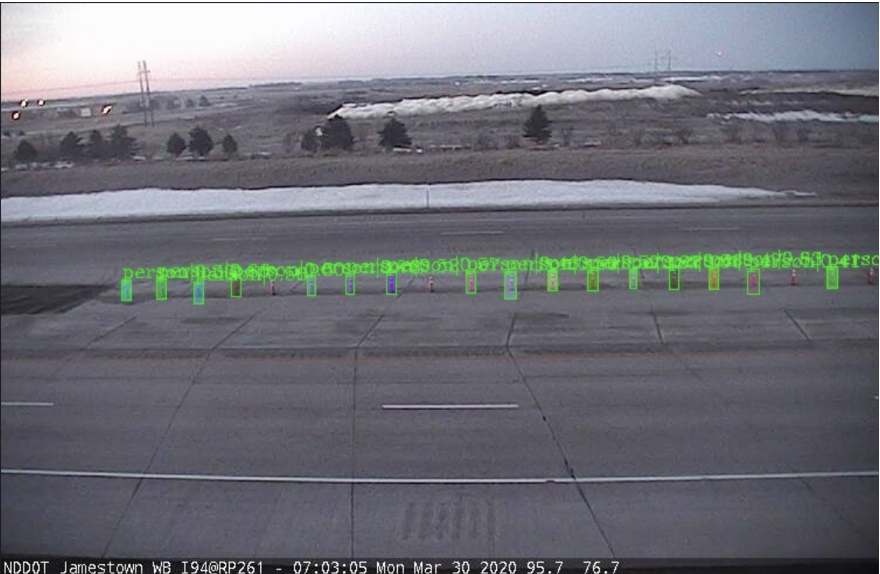}}
    \subfigure[]{
    \includegraphics[width=0.22\textwidth]{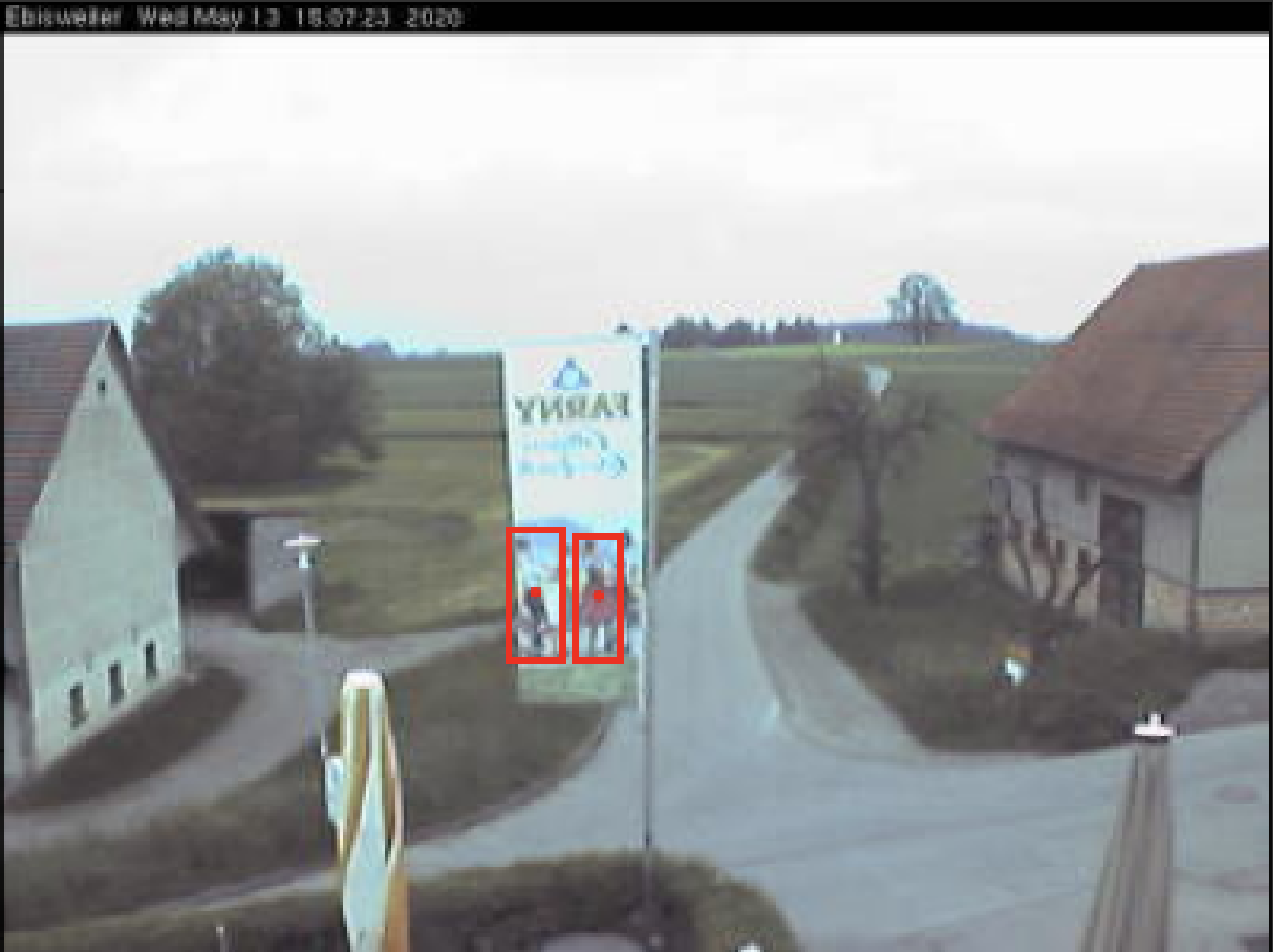}}
    \caption{Failures of the chosen object detection method on selected images. (a) False positives detected on a road in Jamestown, ND, USA. (b) 2 people detected incorrectly in an image from a banner in Badneuenahr-ahrweiler, Germany. Although to human eyes it is obvious these people are drawn on a banner, object detection does not successfully take into account the context. Image sources: ND DOT, Webcamgalore.com}
    \label{fig:failures}
\end{figure}

\subsection{Calculating Group Sizes}
     To calculate group size, we create an unweighted graph in which all of the detected persons are each represented by a node, and edges are drawn between nodes based on social distancing violation. For example, if two people are within six feet of each other, we draw an edge between their two nodes. We use two definitions of a group to create a lower bound and upper bound of the largest group found in a given image. The lower bound is defined as the largest adjacency list of the graph. The upper bound is defined as the largest connected component of the graph. Since we are not aware of a concrete definition of what constitutes a `group' of people, we use these boundaries to estimate group sizes.

\section{Results}
Here we discuss successes and failures of the computer vision models chosen, the presented social distancing algorithm, crowd and group size statistics, violation statistics, and worldwide trends over time.

\begin{figure*}
    \centering
    \includegraphics[width=0.65\textwidth]{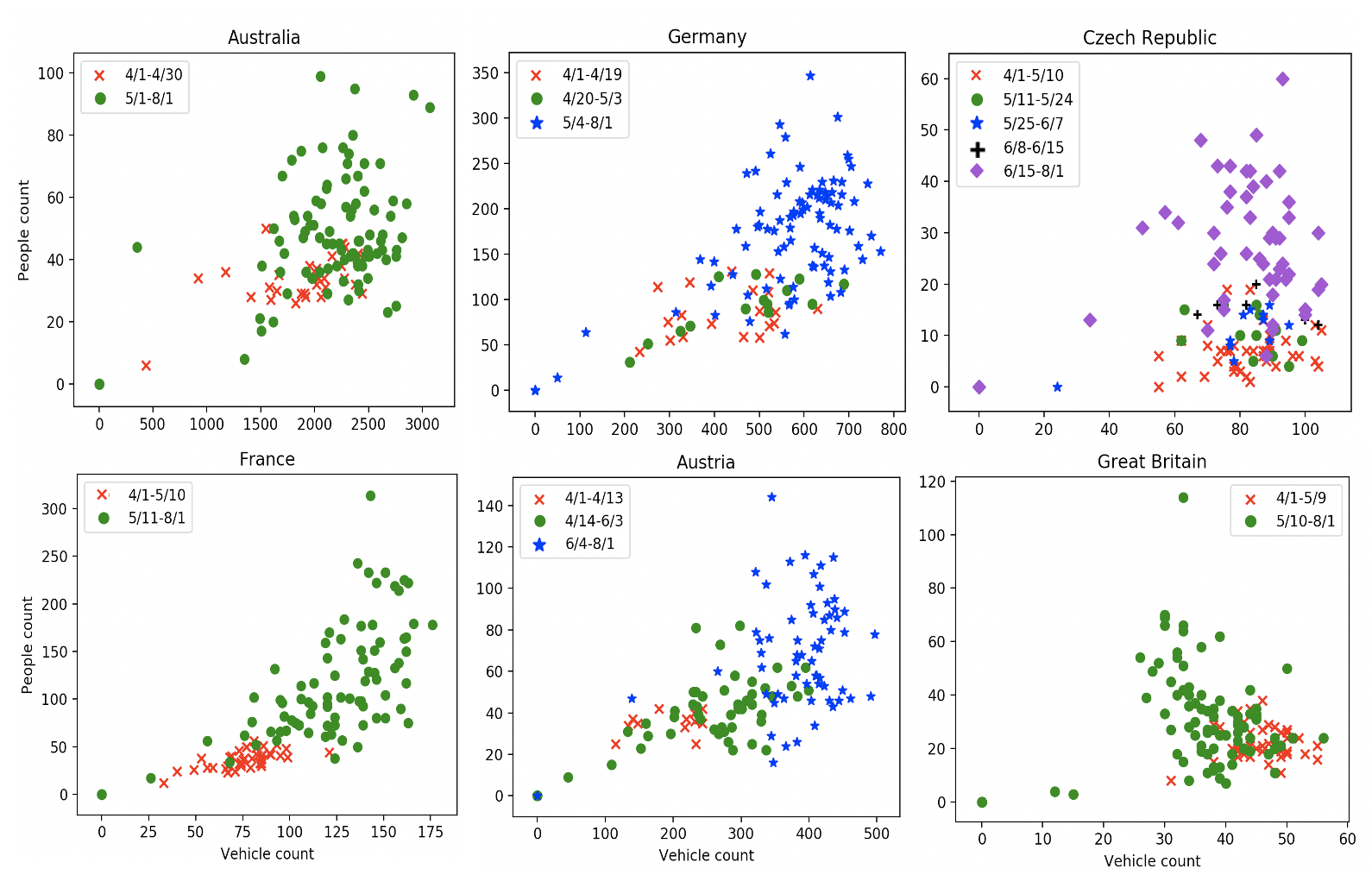}
    \caption{Scatterplots of daily data from select countries. Different colors and markers represent different date ranges indicated in the legend of each plot to reflect the period of lockdown and reopening. Red crosses represent lockdown. Other symbols represent different phases of reopening.}
    
    \label{fig:scatter_countries}
\end{figure*}

\begin{figure*}
    \centering
    \includegraphics[width=\textwidth]{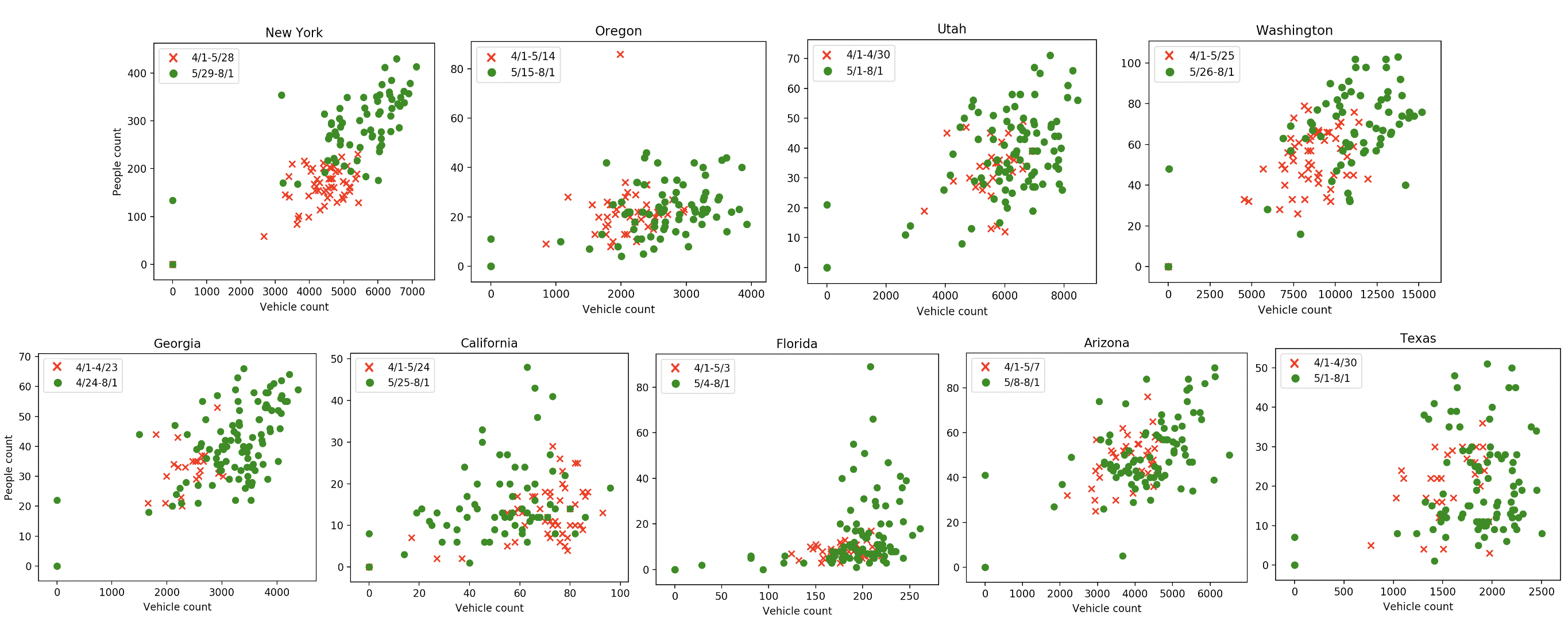}
    \caption{Scatterplots of daily data from select US states. Date ranges indicated in legend. Red crosses represent lockdown and green circles represent reopening.}
    \label{fig:scatter_states}
\end{figure*}

\vspace{-1mm}

\subsection{Performance of Computer Vision}

We show some successes of computer vision on worldwide network camera data. We do not show the bounding boxes on the following images as these would impede the reader's view of the people. Annotated images are available upon request.
Figure~\ref{fig:successes} (a) shows 35 people detected in  Ocean Beach, Mississippi. Although most people occupy very few pixels in the image, the model gives us a relatively accurate estimate. More than 80 people are detected in Figure~\ref{fig:successes} (b) during a protest. Although we do not utilize crowd density methods, object detection still gives an estimate useful enough to allow observation of changes over time.

Given the diversity and realistic nature of our data, we find some limitations. For example, Figure~\ref{fig:failures} (a) shows a road with separators detected as people. Figure~\ref{fig:failures} (b) shows an advertisement on a banner showing people. The object detector consistently detects 2 people because object detection does not take into account context. These results could be ameliorated by increasing the confidence threshold of our detector. However, we find that in many cases, the resolution of data is so low that many true positives are detected with low confidence threshold between 0.3 and 0.4. Thus, there is a trade-off to be considered when using a static object detection model on image data at this scale.

\subsection{Social distancing violations}

\begin{figure}
    \centering
    \subfigure[]{\includegraphics[width=0.22\textwidth,height=3cm]{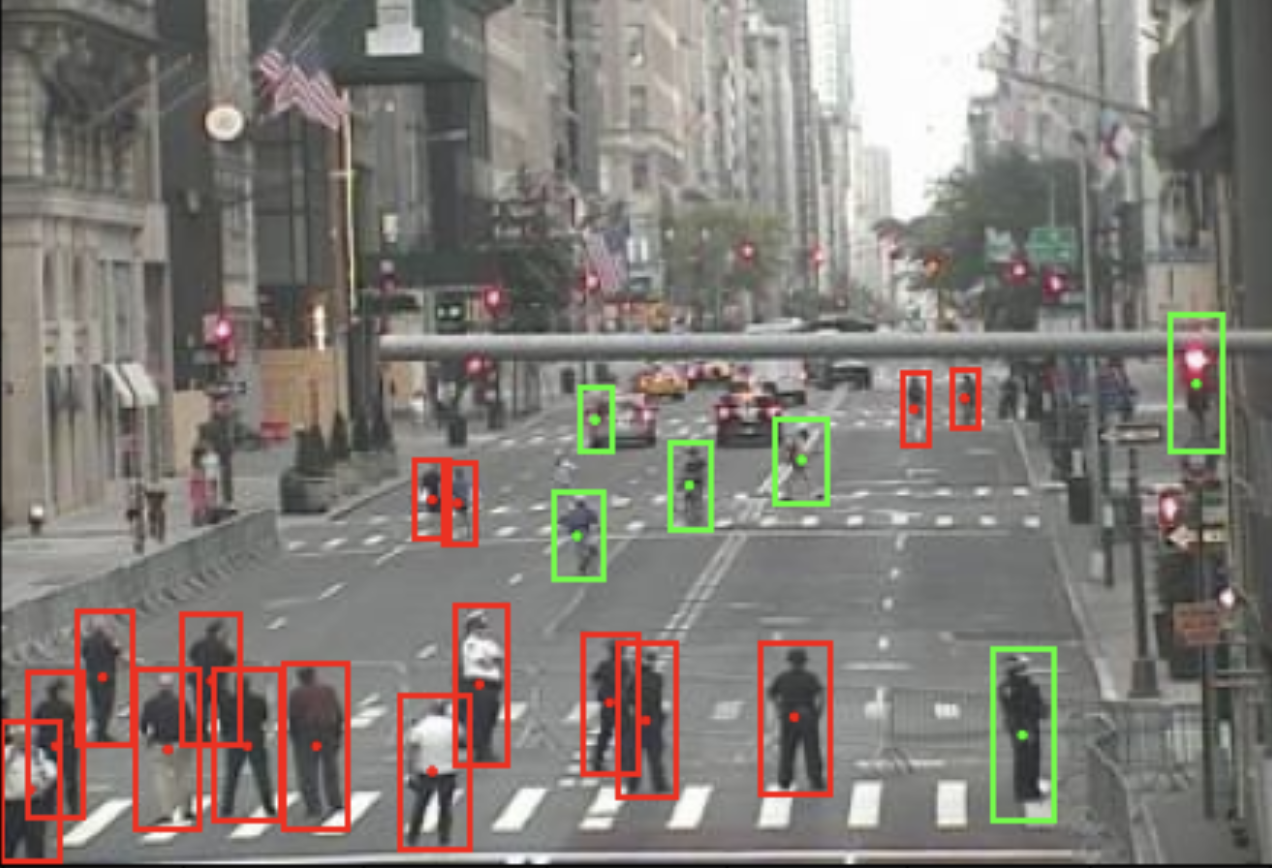}}
    \subfigure[]{\includegraphics[width=0.22\textwidth,height=3cm]{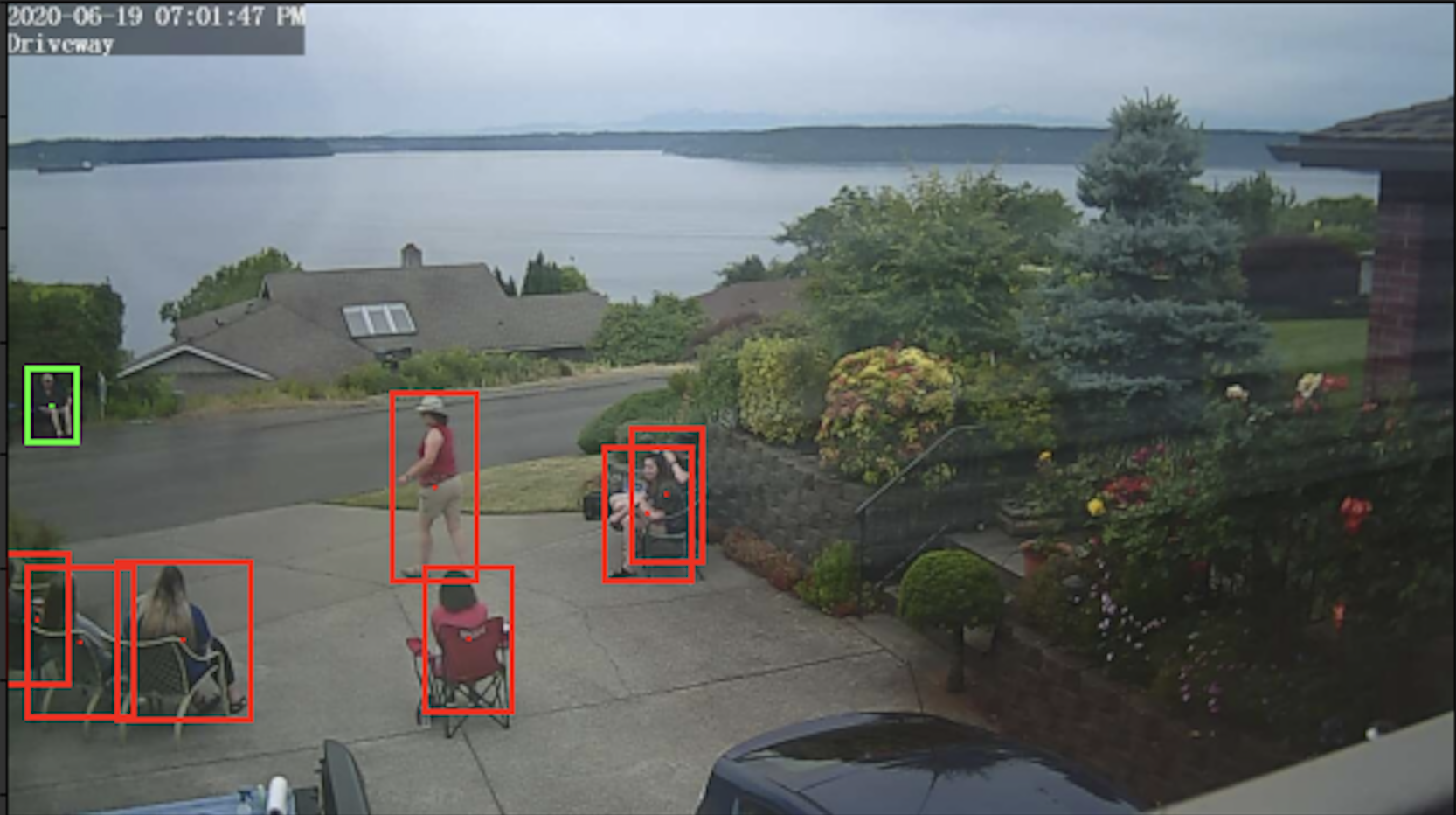}}
    \caption{Examples of the proposed algorithm performing in situations with varying camera angle and depth differences between bounding box detections. (a) A street in New York City, NY, USA. Image source: NY DOT. (b) Backyard in Tacoma, WA, USA. Image source: Wunderground.com}
    \label{fig:12}
\end{figure}

\vspace{-2pt}
We utilize our social distancing algorithm to find specific locations in the world exhibiting low levels of social distancing. We show two examples here. 
Figure~\ref{fig:12} (a) shows  police officers standing more closely together than pedestrians on the street in New York City, USA. Figure~\ref{fig:12} (b) shows people gathering in a backyard in Tacoma, WA, USA. A full list of images found using this method can be accessed at the Zenodo upload \cite{isha_ghodgaonkar_2020_3996718}.

\begin{figure}
    \centering
    \includegraphics[width=0.45\textwidth]{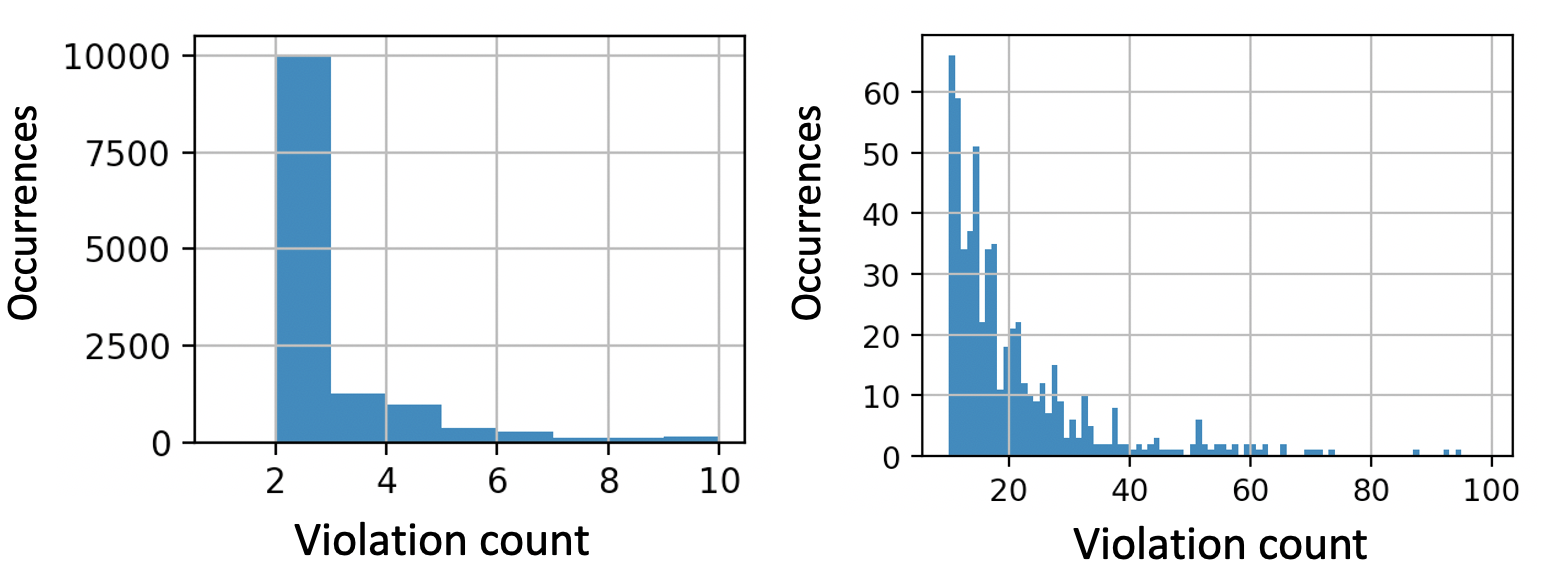}
    \caption{Histogram of people counts observed over all images. Left: zoomed in counts of 1-10 people. Right: Counts of 11-100 people. 0 people are detected more than 1,200,000 times and are not shown.}
    
    \label{fig:ped}
\end{figure}

\begin{figure}
    \centering
    \includegraphics[width=0.45\textwidth]{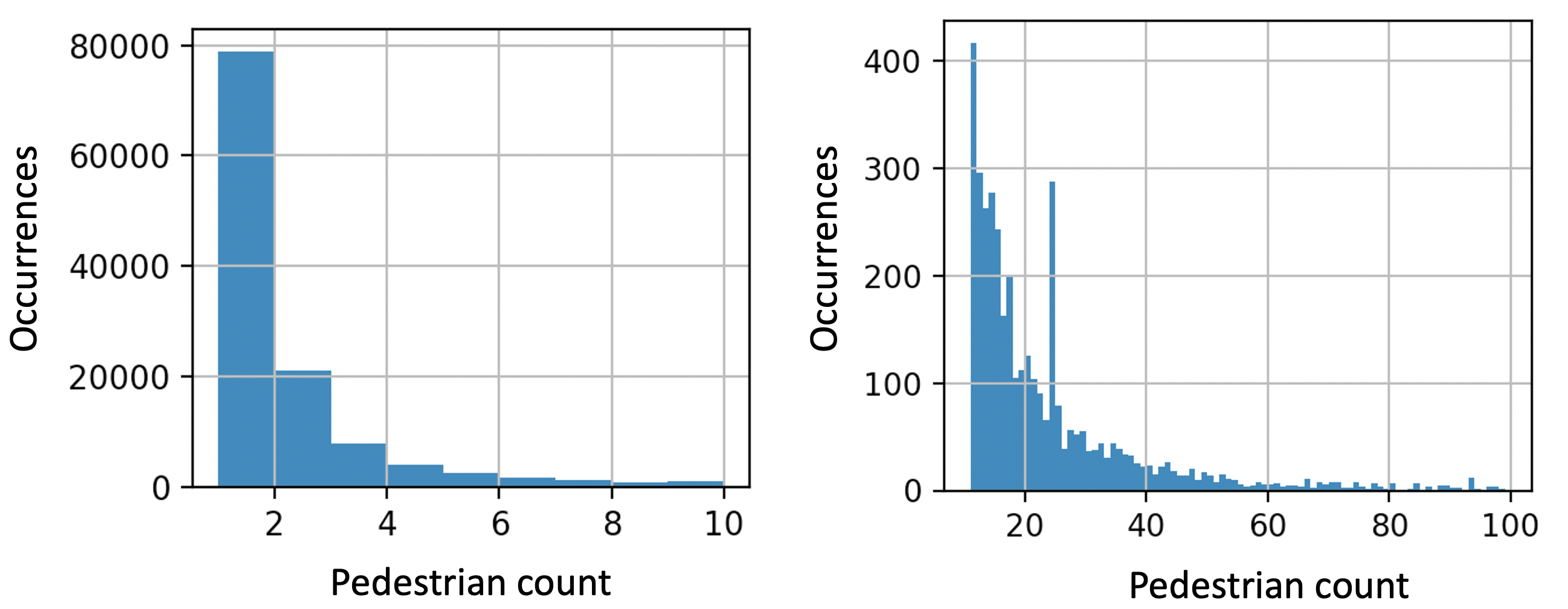}
    \caption{Histogram of violation counts observed over all images. Left: zoomed in counts of 1-10 people. Right: Counts of 11-100 people. 1 violation is not possible as 2 people must be in violation with each other. More than 1,000,000 images have no violation and are not shown.
    }
    
    \label{fig:vio}
\end{figure}

\begin{figure}
    \centering
    \includegraphics[width=0.45\textwidth]{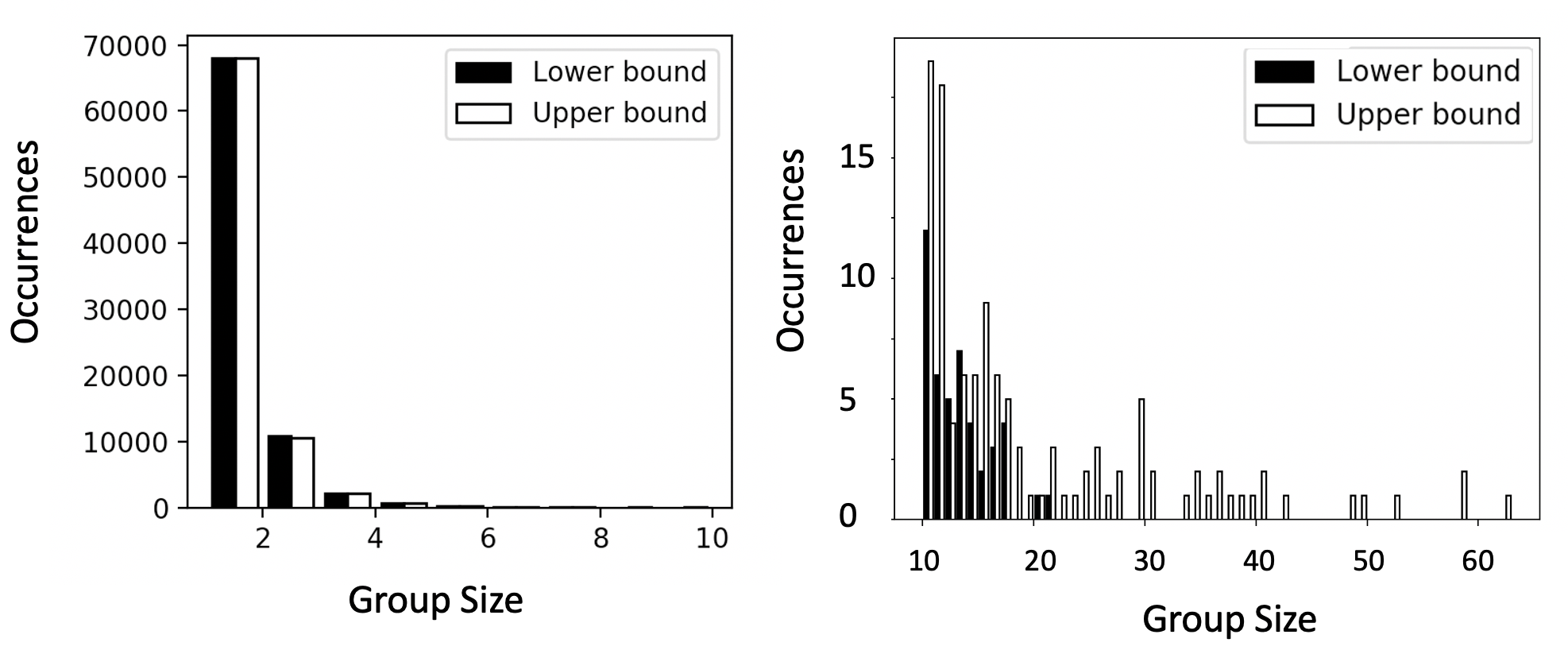}
    \caption{Histogram of group counts observed over all images. Left: Group counts of 1-10 people. Right: Group counts of 11-100 people. Lower bound and upper bound are calculated as described in VI-F.
    }
    
    \label{fig:group}
\end{figure}

\subsection{Trends in Crowd Sizes}
We observe pedestrian counts of up to 100, given the limitation of the model chosen. We observe the distribution of pedestrian counts over all images as shown in Figure~\ref{fig:ped}. We observe a similar distribution of violation counts as shown in Figure~\ref{fig:vio}. We detect group sizes up to 63 using the loose definition (upper bound), and sizes of up to 21 using the strict definition (lower bound), as shown in Figure~\ref{fig:group}. The data shows that big groups are occasionally detected. Smaller group sizes and pedestrian counts are more frequent than larger ones.

\subsection{Observed Trends in Select Countries}
\begin{figure*}
    \centering
    \includegraphics[width=\textwidth]{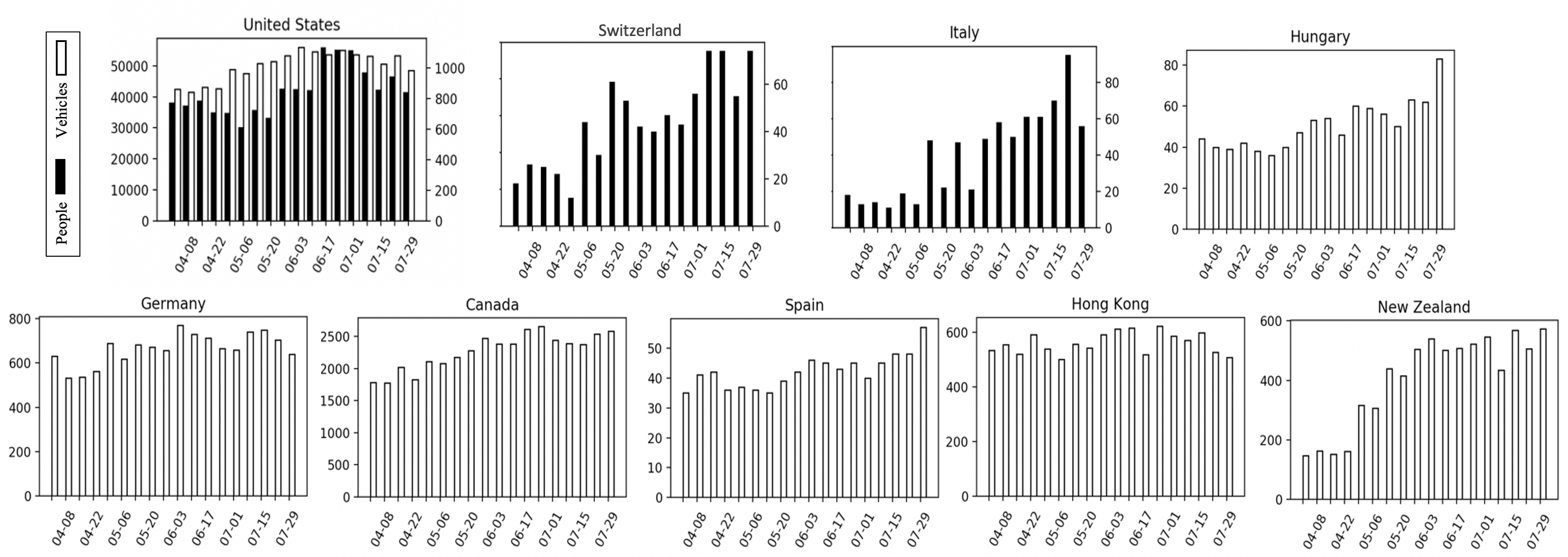}
    \caption{Vehicle and people counts from April 1 to August 1. Weekly maximums are shown from the period starting with the date indicated on the x-axis and ending with the date previous to the next date on the x-axis.}
    \label{fig:plots}
\end{figure*}

Figure~\ref{fig:plots} shows the numbers of people and vehicles over time detected by cameras in 9 countries. We show people data for which the maximum number of people detected is at least 40 and vehicle data for which the maximum number of vehicles detected is at least 50. We do not set a minimum threshold because we expect to observe fewer people during lockdown, and thus do not rule out days where the number of people detected is minimal. In some countries with relatively few cameras, we still observe trends. It is important to note that these trends may not be representative, but may still provide insight. Figure~\ref{fig:scatter_countries} shows scatter plots of people counts vs. vehicle counts in 8 countries, where we do not filter any data. The (0,0) points on all charts are due to interruptions in data collection. Here we discuss trends in a few countries of note with sufficient data.

\textbf{United States}
In the United States, we observe a slight increase in the number of vehicles on roads over time and an unclear trend in the numbers of people in public over time as shown in Figure~\ref{fig:plots}. Social distancing policies in the United States vary by state. As of March 31, 30 states published stay at home orders, and as of April 12, 12 more states did the same \cite{lee_mervosh_avila_harvey_matthews_2020}. Reopening plans also vary by state, ranging from late April to early June. We do not observe a clear response to these orders in the data on a national scale. We show a few states of interest in Figure~\ref{fig:scatter_states}. We see that after respective reopening on April 24 \cite{lee_mervosh_avila_harvey_matthews_2020} in Georgia, we see a slight increase in activity. We see a clearer increase in activity after New York's June 8 reopening \cite{the_new_york_times_2020}. We see a slight increase in activity after May 26 in Washington. California shows a decrease in vehicle count and an increase in people count after May 25. We see an increase in people count following May 4 in Florida. Oregon shows almost no change after May 15, as do Utah and Texas after May 1, and Arizona after May 8.

\textbf{France}
On March 16, France announced mandatory home confinement, a policy which was extended until May 11 \cite{magnenou_matalon_2020}. After May 11, a clear increase in activity can be observed.

\textbf{Germany}
On March 22, Germany banned groups of 2 or more people in public \cite{welt_2020}. Small businesses were reopened on April 20 \cite{connolly_2020}, and on May 4, some schools and other institutions reopened \cite{www_dw_com}. We observe a general increase of activity in public areas following these dates.

\textbf{Australia}
In Australia several relaxations of restrictions occurred between April 27 and May 2 by various regions and states \cite{digital_story_innovation_team_by_inga_ting_and_alex_palmer_2020}. We show the changes happening after May 1 and observe an increase in activity. 

\textbf{Austria}
Austria reopened small shops on April 14 \cite{orf_at_2020}, and lifted border restrictions on June 4 \cite{gotev_2020}. We observe a steady increase in the numbers of both people and vehicles in public throughout these time periods.

\textbf{Great Britain}
On May 10, Great Britain changed their motto from ``stay at home" to ``stay alert" \cite{gov_uk}. We observe an increase in people count but a decrease in vehicle count following this date.

\textbf{Czech Republic}
The Czech Republic opened museums and cinemas May 11, restaurants and bars May 25, increased public gathering size on June 8, and opened borders on June 15 for select countries from the EU \cite{measures_adopted_czech_government}. We observe that vehicle counts do not show much change over time but people counts increase steadily.

\vspace{1mm}
\section{Discussion}
Here we discuss limitations of the work presented and potential future directions of research.

\subsection{Performance of Computer Vision}
While the object detection models utilized represent the state-of-the-art in computer vision, data obtained from cameras are diverse and thus pose challenges. There are cases of false positives and negatives for particularly difficult cases and low quality data. While the chosen methods can provide reasonable estimates of numbers of people and vehicles in images resulting in the ability to observe trends, false detections may pose a problem for quantifying social distancing as the proposed method depends upon accurate bounding boxes. To avoid compounding the error, even more robust object detection models should be developed. 

\subsection{Calculating Social Distancing and Group Sizes}
We observe variations in bounding box area and aspect ratios leading to misclassifications. For our purposes these misclassifications are negligible, but improving the method is a direction of future work, such as by using Disnet \cite{haseeb2018disnet}.
Additionally, this study does not consider groups of individuals that may belong to the same family or residence. Therefore, the number of violations in reality may be lower than the ones reported here, although the purpose of our documentation of violations is simply to show trends. We additionally attempt to detect group sizes using two separate definitions based on graph theory and display distributions of the group sizes detected. However, once the groups are identified, it is impossible to tell whether the individuals in a given cluster belong to the same residence or not without invading privacy. Graph theory also removes the spatial dimension of data which is useful for determining group sizes based on spatial heuristics. Therefore, definitively measuring social distancing is an emerging area of research.

\subsection{Observed Worldwide Trends}

While this work provides insight into responses to social distancing guidelines, trends may not be representative of each geographical location. The results presented here are limited to locations observed by our camera network. More representative results could be obtained with a more expansive camera network, which is a direction of future research. For example, the trends observed in Great Britain and California do not resemble the trends observed in other US states or European countries.
\vspace{-1mm}

\vspace{1mm}
\section{Conclusion}

This paper seeks to answer the following questions: In the time of COVID-19, to what extent can computer vision analyze worldwide network camera data for observing human activities? This paper presents the collection of live visual data from worldwide network cameras for monitoring responses to social distancing policies, successes and failures of powerful object detection models to detect people and vehicles in this diverse data, a social distancing algorithm, and trends observed in 15 countries. We find that existing vision methods perform well enough on challenging data to allow us to observe preliminary trends, but are not perfect. We utilize these methods to collect and display aggregate numbers of people and vehicles over time in select countries. We show the potential to observe responses to social distancing guidelines using network cameras. 
Readers interested in obtaining the data collected and full analysis may contact the project's lead author, Isha Ghodgaonkar (ighodgao@purdue.edu).

\section*{Acknowledgments}
\vspace{-1mm}
This research uses resources of the Argonne Leadership Computing Facility, which is a DOE Office of Science User Facility supported under Contract DE-AC02-06CH11357. We thank the Argonne Leadership Computing Facility for access to the Cooley supercomputer that was used in this study. This project is supported in part by the National Science Foundation OAC-2027524 and Purdue College of Engineering Fund for Security and Privacy.  Any opinions, findings, conclusions or recommendations presented in this material are those of the authors and do not necessarily reflect the views of the National Science Foundation.

\def\bibfont{\footnotesize}
\bibliographystyle{abbrv}
\bibliography{references/refs,references/caleb-zotero-refs_edited}

\end{document}